# Prediction of the Yield of Enzymatic Synthesis of Betulinic Acid Ester Using Artificial Neural Networks and Support Vector Machine

Run Wang[1], Qiaoli Mo[2], Qian Zhang[1], Fudi Chen[3] and Dazuo Yang[3, 4,*]

[1]College of Light Industry, Textile and Food Science Engineering, Sichuan University, Chengdu, Sichuan 610064, China

[2]College of Chemistry, Sichuan University, Chengdu, Sichuan 610064, China

[3]Key Laboratory of Marine Bio-Resources Restoration and Habitat Reparation in Liaoning Province, Dalian Ocean University, Dalian 116023, China

[4]College of Life Science and Technology, Dalian University of Technology, Dalian 116021, China

[*]Email: dzyang1979@hotmail.com

Tel: +86-0411-84762290

Fax: +86-0411-84762290

**Abstract**

3β-*O*-phthalic ester of betulinic acid is of great importance in anticancer studies. However, the optimization of its reaction conditions requires a large number of experimental works. To simplify the number of times of optimization in experimental works, here, we use artificial neural network (ANN) and support vector machine (SVM) models for the prediction of yields of 3β-*O*-phthalic ester of betulinic acid synthesized by betulinic acid and phthalic anhydride using lipase as biocatalyst. General regression neural network (GRNN), multilayer feed-forward neural network (MLFN) and the SVM models were trained based on experimental data. Four indicators were set as independent variables, including time (h), temperature (℃), amount of enzyme (mg) and molar ratio, while the yield of the 3β-*O*-phthalic ester of betulinic acid was set as the dependent variable. Results show that the GRNN and SVM models have the best prediction results during the testing process, with comparatively low RMS errors (4.01 and 4.23respectively) and short training times (both 1s). The prediction accuracy of the GRNN and SVM are both 100% in testing process, under the tolerance of 30%.

**Introduction**

3β-*O*-phthalic ester of betulinic acid has clinical potential as an anticancer medicine, which can be synthesized from reaction of betulinic acid and phthalic anhydride using lipase as biocatalyst (Fig. 1). It has a variety of properties including inhibition of antibacterial, anti-inflammatory, anti-malarial, anthelmintic, antioxidant and human immunodeficiency virus (HIV) (Yogeeswari, 2005).According to previous studies, the introduction of polar groups at the C-3 and C-28 positions also highly increases the anticancer activity and hydro-solubility (Thibeault *et al*., 2007; Gauthier and Legault, 2008).However, the practical applications of betulinic acid in the pharmaceutical and medical industry is deeply constrained because it is insoluble in water (approximately 0.02 mg/mL) under ordinary circumstances, leading to great difficulties in preparation of injectable formulations for biological experiments and decreases the bioavailability.

The detailed approaches for the synthesis of 3β-*O*-phthalic ester of betulinic acid based on chemical catalytic esterification have been reported by previous research reports(Mukherjee *et al*., 2004; Kvasnica *et al*., 2005; Mukherjee *et al*., 2006; Rajendran *et al*., 2008), which have several disadvantages (e.g. high energy consumption and by-products) (Yasin *et al*., 2008). Compared with traditional chemical approaches, the application of enzymes in organic synthesis offers a series of advantages, including high catalytic efficiency, high selectivity, mild reaction condition and high product purity and quality (Loughlin, 2000; Zarevúcka and Wimmer, 2008). However, the best detailed conditions for the synthesis are difficult to obtain due to the large-scale and complex laboratory experiments. Moghaddam and colleagues(2010) used artificial neural network (ANN) models to develop models for predicting the yield of enzymatic synthesis of betulinic acid ester. They successfully found that the quick propagation algorithm was the best model during their computational experiments. Nevertheless, previous ANN models are based on comparatively complex operations and the selection method of the best ANN model was based on a limited number of results, which are not robust and user-friendly enough, compared with latest machine learning models. Here, we aim at using novel and user-friendly approaches of ANN models and support vector machine (SVM) to train the data of the yield of enzymatic synthesis of betulinic acid ester, and obtain a series of best machine learning models for the prediction of the yield. Comparisons are made in order to determine the most suitable machine learning model for the prediction.

## 2 Materials and Methods

2.1 Data Set

According to previous research, the synthetic conditions of enzymatic synthesis of betulinic acid ester includes time (h), temperature (℃), amount of enzyme (mg) and molar ratio(mmol betulinic acid/mmol phthalic anhydride) (Moghaddam *et al*., 2010). Here, we aim at using novel ANN and SVM models to fit the four conditions and to predict the isolated yield (%) of the enzymatic synthesis.

A complete machine learning model consists of two parts, the independent variable(s) and the dependent variable(s). Here, we set the time (h), temperature (℃), amount of enzyme (mg) and molar ratio as independent variables, while the isolated yield (%)was set as the dependent variable. 65% data group was set as training set, which 35% data group was set as testing set.

2.2 ANN models

A series of statistical learning algorithms with the name of Artificial Neural Networks (ANN) could give judgments when inputted into a large amount of information (Hopfield, 1988; Judith and DeLeo, 2001; Yegnanarayana, 2009). Like the brain, a biological neural network, ANNs are usually comprised of neurons that can make instant calculations in different conditions with the connection with each other. Different from ordinary networks with one or two layers, there are three layers in Artificial Neural Networks which can learn inputs efficiently and recognize patterns in a direct way. Furthermore, ANNs can also use complicated algorithms to make prediction and find the optimum solution. Therefore, when dealing with problems that are too complex to solve, ANNs can take the place of human brains, and the application of ANNs is more and more popular in the scientific research. In this passage, we introduce the use of two kinds of ANNs, multilayer feed-forward neural networks (MLFN) and general regression neural networks (GRNN) to build the models to forecast the yield of enzymatic synthesis of betulinic acid ester.

1.2.1 Multilayer feed-forward neural networks (MLFN)

With the training of a back-propagation learning arithmetic, multilayer feed-forward neural networks, one of the most popular neural networks, can be used to predict a large range of chemical reactions. (Johansson *et al*., 1991; Smits *et al*., 1994; Svozil and Kvasnickab, 1997)

Neurons in the MLFN models are put into different layers (Figure2). Input layer is the first layer and output layer is the last one. Between them, hidden layers play a role of calculating and modeling. To be specific, we could use the mapping function $\Gamma$ that allocates a subset $\Gamma(i) \subseteq V$ to each neuron $i$ to describe the neurons in a formal way, and The subset $\Gamma(i)$ is made up of all ancestors of the neuron. Meanwhile, there is a subset $\Gamma(i)^{-1} \subseteq V$ containing all ancestors of the given neuron $i$. All neurons in a given layer is connected with any one of the neurons in the past layer. A weight coefficient $\omega_{ij}$ can be used to present the connection of the $i$th and $j$th neuron, and we apply the threshold coefficient $\vartheta_i$ (Fig. 3) to present the $i$th neuron. The level of significance of a particular connection in the neural network can be indicated by the weight coefficient. Additionally, Eqs.(1) and (2) can determine the output value (activity) of the $i$th neuron $x_i$. It holds that:

$$x_i = f(\xi_i) \tag{1}$$

$$\xi_i = \vartheta_i + \sum_{j \in r_i^{-1}} \omega_{ij} x_j \tag{2}$$

Where the potential of the $i$ th neuron is represented by $\zeta_i$ and function $f(\zeta_i)$ indicates the transfer function (the summation in Eq. (2) runs over all neurons $j$ devolving the signal to the $i$th neuron). The threshold coefficient could be comprehend as a weight coefficient of the connection with regularly added neuron $j$, where $x_j = 1$ (so-called bias).

For the transfer function it holds that:

$$f(\zeta) = \frac{1}{1+exp(-\zeta)} \tag{3}$$

To minimize the total of the squared differences between the required and calculated output values, The weight coefficients $\omega_{ij}$ and threshold coefficients $\vartheta_i$ are devolved by the supervised adaptation process. Minimization of the objective function E can complete this by:

$$E = \sum_0 \frac{1}{2}(x_0 - \hat{x}_0)^2 \tag{4}$$

Where $x_0$ and $\hat{x}_0$ are vectors comprised of the required and calculated processes of the output neurons and the summation carried out over all output neurons $o$.

1.2.2 General regression neural network (GRNN)

The Nadaraya-Watson kernel based general regression neural network (GRNN) which is put forward by Specht (1991) is widely applied in medical diagnosis, pattern identification, forecasting, three-dimensional modeling, chemical engineering and function approximation (Hoskins and Himmelblau, 1988; Khan *et al*., 2001; Goulermas *et al*., 2007; Kandirmaz *et al*., 2014; Li and Leng *et al*., 2014; Li and Wang *et al*., 2014). Compared with other statistical neural networks like feed-forward networks, GRNN represents more accurate in regard to function approximation (Kandirmaz *et al*., 2014). Even if it is the first time to be used with the aim of function approximation, in some studies, GRNN is also applied to classification problems with small modifications (Kandirmaz *et al*., 2014). As can be seen from Figure 4, there are 4 layers in GRNN, respectively, input layer, pattern layer, summation layer, and output layer, with the characteristics of rapid learning, coherence, and finding optimum with a great many specimens(Yang and Li, 2014).

In the input layer, corresponding inputs are conserved and every input vector $x$ can be devolved to pattern layer which contains different neurons for training data. As illustrated in Eq. (5), calculations are made about weighted squared Euclidean distance in the pattern layer. Before aggregated, inputs which are used for activation function, whether squares or absolute values, should be deducted from neuron values in pattern layer when applying to network. Then, neurons in summation layer, to which the outcomes are transferred, add dot product of pattern layer weights and outputs. As can be seen from Figure 4, $f(x)K$ indicates weighted outputs of pattern layer where $K$ is a Parzen window related constant. $Yf(x)K$ refers to multiplication of training data output $Y$ values and pattern layer outputs. In output layer, $f(x)K$ separates $Yf(x)K$ to estimate desired values, held in Eqs. (6) and(7) (Goulermas *et al*., 2007; Yang and Li, 2014):

$$D_j = (x - x_j)^T (x - x_j), (5)$$

$$Y(x) = \frac{\int_{-\infty}^{\infty} Yf(x,Y)dY}{\int_{-\infty}^{\infty} Yf(x,Y)dY}, (6)$$

$$Y(x) = \frac{\sum_{j=1}^{p} y_j e^{[-D_j/2\sigma^2]}}{\sum_{j}^{p} e^{[-D_j/2\sigma^2]}} \quad (7)$$

2.3 SVM model

There is a formidable machine learning technique with the name of support vector machine (SVM) established from the statistical learning theory (Deng *et al*., 2012). In terms of increasing generalization, this theory can give an integral optimization in an efficient way, with restricted information of specimens between the learning capacity and the complicacy of models. Separating all specimens with the maximum margin, the main theory of SVM is a plane which has the ability of discovering the optimal hyperplane and linear separable dualistic classification (Zhong *et al*., 2013; Chen *et al*., 2015). Additionally, the plane can also increase the forecasting ability of the model and can decrease the mistake occurring accidentally when classifying. As can be seen from Figure 5, specimens of type 1 are represented by "+" and specimens of type −1 are represented by "−" to shows the optimal hyperplane.

To explain the main structure of a representative support vector machine, Figure 6 shows a small subset derived from the training data by related algorithm that contains the SVM. Kernels are characterized by the letter "*K*" (Kim *et al*., 2005). To have a forecasting accuracy, appropriate kernels and suitable parameters should be selected in terms of classification. However, we could not find an available international standard to select these parameters. In most cases, to solve this task in a relatively reasonable way, we could take the advantage of the experiences from massive calculations, the contrast of experiment results, and the application of cross validation which is realizable in program package (Fan *et al*., 2008; Guoand Liu, 2010; Chen *et al*., 2015).

2.4 Model Development

The ANN prediction models were constructed by the NeuralTools® software (trial version, Palisade Corporation, NY, USA) (Pollar and Jaroensutasinee, 2007; Friesen *et al*., 2011; Vouk *et al*., 2011). The GRNN and MLFN was chosen as the learning machines of ANNs.

We used RMS error and training time as the indicators to measure the performances of ANN and SVM models (Table 1). The nodes of MLFN models were set from 2 to 25, from which we could find out the change regulation of the MLFN models when dealing with development processes.

Table 1 indicates that the GRNN, SVM and MLFN with 4 nodes have comparatively low mean RMS errors (4.01, 4.23and 5.56 respectively). It is clear that the GRNN and SVM have the lowest RMS errors and training times, while the MLFN models have comparatively higher RMS errors and longer training times. In addition, the prediction accuracy (under the tolerance of 30%) of the GRNN and SVM are both 100%. Here, we discuss the availability of the GRNN, SVM and MLFN respectively in order to determine the most suitable model for the prediction of the yield.

## 3 Results and Discussion

3.1 Comparison between the GRNN and MLFN

As for the GRNN, it has the lowest RMS error and training time during our research, compared with other 24 MLFN models. And according to the robustness of the principles of the GRNN, it has a high reproducibility, which has an overwhelming advantage compared with other ANN models during our research. In order to test the robustness of the GRNN, the computational experiments for the GRNN was repeated, which are shown in Figure 7:

Figure 7 shows the RMS errors of the GRNN models in repeated experiments. It is significant that there is a stable fluctuation during repeated experiments, which shows that the GRNN model for the optimization process is robust. And what is more importantly, the mean RMS error is relatively low, which ensures the availability of the GRNN model.

To illustrate the change of the MLFN models with different numbers of nodes, Figure 8 is used for showing the results of MLFN models with different nodes.

It can be seen that with the increase of nodes, the RMS errors and training time of MLFN models become unsteadily fluctuant, which highly corresponds to the fluctuation character of typical MLFN models. It is worth mentioning that results in different MLFN models presented by Table 1 is not a fixed result, because of the effects of different random initial values chosen by the computer when training. However, it is still clear that the MLFN model may have a good result (relatively low RMS error and short training time) in a relatively low number of nodes. For practical applications, operators can use related software to find out the most suitable model for the optimization of reaction conditions in the range of low number of nodes. However, compared to the GRNN, MLFN models commonly cost longer training time and the fluctuations are not as stable as those of GRNN model. Therefore, we still consider the GRNN model is a more suitable model for the prediction of the yield of enzymatic synthesis of betulinic acid ester

3.2 Training and Testing Results of the GRNN and the SVM

Here, we use one of the typical examples of the training and testing results to present the availability of the GRNN, and also illustrate the testing results of the SVM. Figures 9 and 10 are used for the illustration of the training and testing results of the GRNN, while Figure 11 is used for the illustration of the testing results of the SVM. The training and testing sets of the GRNN and SVM are the same.

For showing the capacity for recall of the GRNN model for the optimization of the design, Figure 9 is used for illustrating the training results of the GRNN.

Figure 9 shows that the GRNN model has a strong capacity for recall. The predicted values is highly close to the actual values (Figure 9. (a)), which indicates that the non-linear fitting effects of the model is highly decent. The comparisons between the residual values and actual/predicted values (Figure 9. (b) and (c)) also show that the residual values are relatively low, which suggests the robustness of the development of the GRNN model.

For showing the availability of the GRNN model after a training process, we use the data set which has not been used for the training process. Results are shown in Figure 10.

Figure 10 shows the precise predicted results during the testing process. Predicted values are close to the actual values [Figure 10. (a)]. Residual values presented by Figure 10. (b) and (c) show that the residual values are relatively low. Results present the robustness and availability of the GRNN model when testing.

In terms of the testing results of the SVM, Figure 11 is illustrated for showing the correctness and robustness of the SVM in the prediction section.

Being similar to the results of the GRNN in the aspects of the RMS error and the training time, the testing results of the SVM are also highly similar to those of the GRNN. We can see that the SVM can generate a fairly analogical and precise result, compared with the testing results of the GRNN.

To make a comparison between the GRNN and SVM, we should firstly note that the initial values for the training process of the GRNN are random, leading to different results in repeated experiments. Compared with the GRNN, the SVM has very good repeatable results due to its principle. Therefore, it seems that the GRNN is not as robust as the SVM. However, results of repeated experiments (Figure 7) show that regardless of those fluctuations of RMS errors, the GRNN is also highly robust because the fluctuations are in a controllable range, which ensures the robustness of the GRNN. In terms of the training time, the GRNN and SVM are too short to find out the differences. However, we should note that the GRNN model can be mainly developed by packed software, while the SVM needs to use the Matlab and finish a series of processes, which requires a higher requirements of computer configuration and comparatively longer time. For a more convenient operation, the GRNN seems more practical than the SVM. Nevertheless, due to the high robustness and repeatability, the SVM should not be neglected in practical applications.

Here, enzymatic synthesis of betulinic acid ester is a typical example for the application of machine learning techniques like ANNs and SVM to the prediction of yields in laboratory experiments. It shows that machine learning techniques have a huge potential applications for the prediction of yields in chemical reactions. And what is more importantly, we can optimize the reaction conditions via the "well-trained" models and predict their yields without trying repeatedly in laboratory experiments.

**Conclusion**

3β-*O*-phthalic ester of betulinic acid is of great importance in clinical research. Here, we successfully find that the application of the ANNs and SVM are useful for the prediction of yields of 3β-*O*-phthalic ester of betulinic acid synthesized by betulinic acid and phthalic anhydride using lipase as biocatalyst. Results show that the GRNN and the SVM model have the best prediction results during the testing process, with comparatively low RMS errors (4.01and 4.23 respectively) and short training times (both 1s). The prediction accuracy of the GRNN and SVM are both 100% in testing process, under the tolerance of 30%. Both the GRNN and SVM have very good repeatability and robustness. Our research successfully show that machine learning techniques like the ANNs and SVM can be used for the prediction of yields and optimization of conditions of traditional organic synthesis. What is more, it is also proved that support vector machine is a novel and strong machine learning tool for related research and applications.

**Conflict of Interests**

The authors declare that they have no conflict of interests.

**Financial support**

This work was funded by the National Marine Public Welfare Research Project (nos. 201305002 and 201305043), National Natural Science Foundation of China (no. 30901107), and the Project of Marine Ecological Restoration Technology Research to the Penglai 19-3 Oil Spill Accident (no. 19-3YJ09).

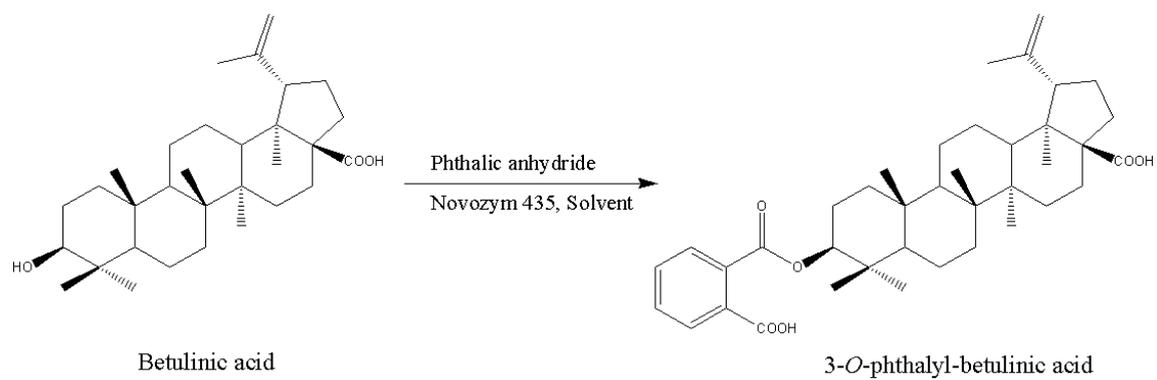

**Figure 1**. Reaction between betulinic acid and phthalic anhydride utilizing Novozym 435 as biocatalyst.

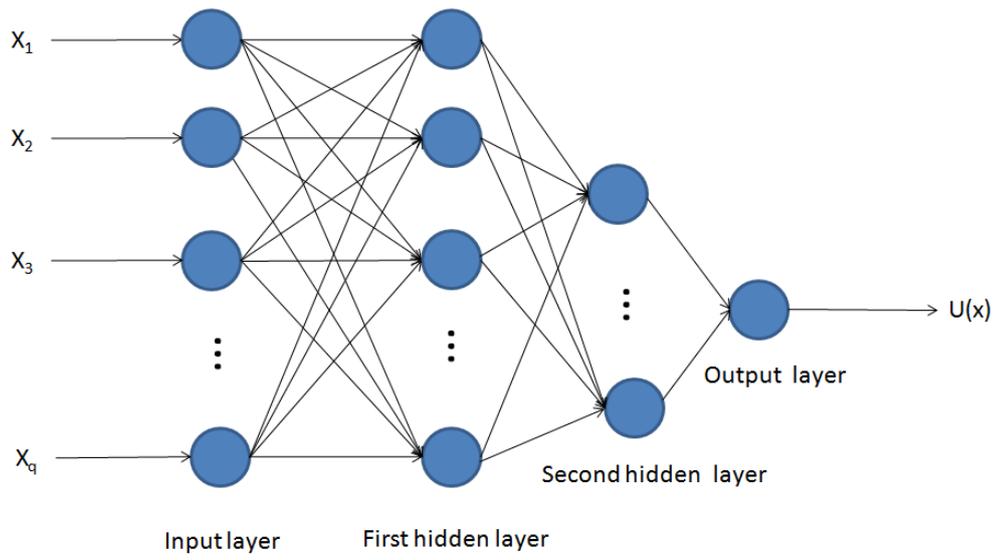

**Figure 2.** Structure of the MLFN.

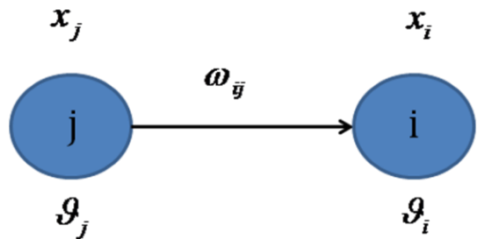

**Figure 3.** Connection between the two neurons $i$ and $j$.

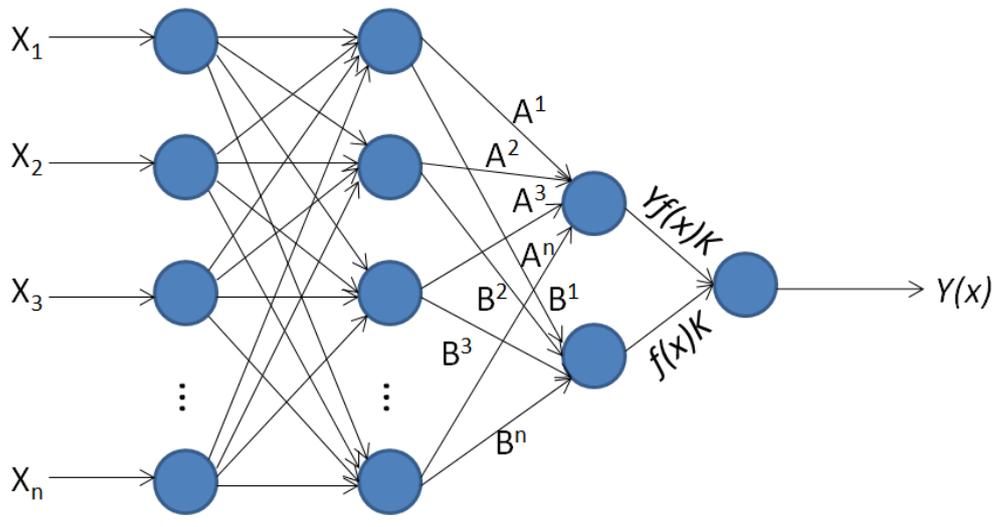

**Figure 4.** Structure of the GRNN.

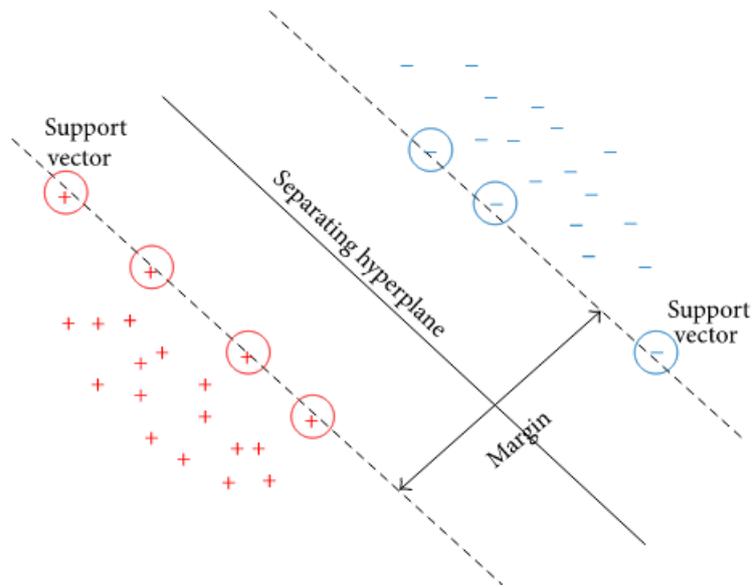

**Figure 5.** The support vectors determine the position of the optimal hyperplane.

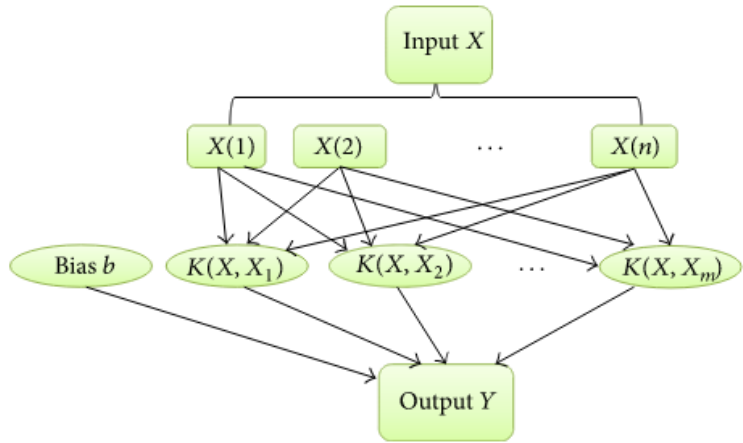

**Figure 6:** The main structure of support vector machine.

Table 1. Best net search in different ANN models.

| Model Type | Mean RMS Error | Training Time | Prediction Accuracy |
| --- | --- | --- | --- |
| GRNN | 4.01 | 0:00:01 | 100% |
| SVM | 4.23 | 0:00:01 | 100% |
| MLFN (2 Nodes) | 6.59 | 0:00:28 | 90.91% |
| MLFN (3 Nodes) | 8.68 | 0:00:29 | 90.91% |
| MLFN (4 Nodes) | 5.56 | 0:00:38 | 90.91% |
| MLFN (5 Nodes) | 7.99 | 0:00:46 | 90.91% |
| MLFN (6 Nodes) | 9.85 | 0:00:57 | 81.82% |
| MLFN (7 Nodes) | 8.98 | 0:01:04 | 81.82% |
| MLFN (8 Nodes) | 10.47 | 0:01:14 | 81.82% |
| MLFN (9 Nodes) | 11.93 | 0:01:26 | 72.73% |
| MLFN (10 Nodes) | 8.17 | 0:01:49 | 90.91% |
| : | : | : | |
| MLFN (25 Nodes) | 54.14 | 0:01:53 | 18.18% |

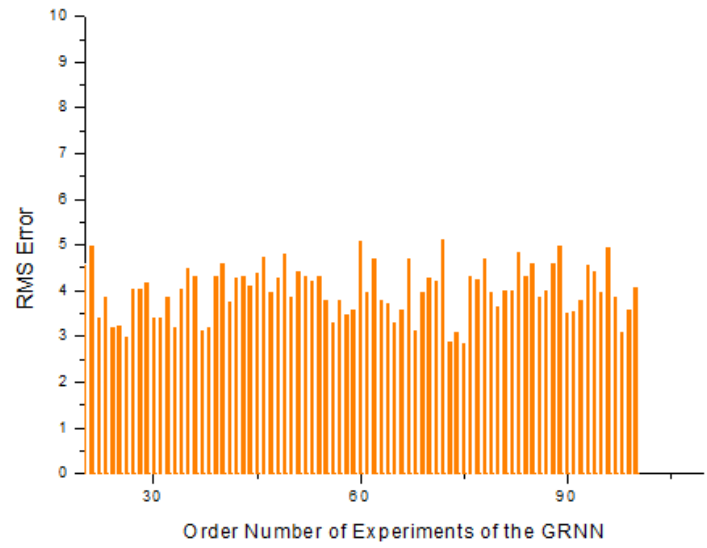

**Figure 7.**Results of computational experiments of the GRNN.

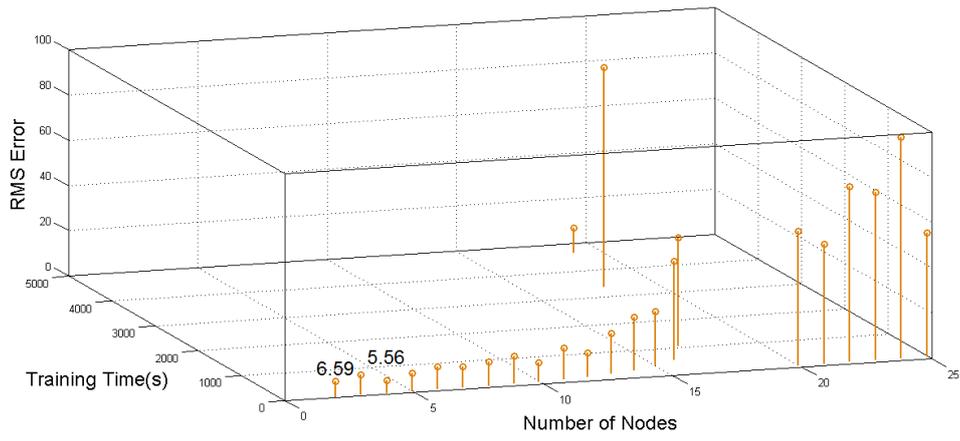

**Figure 8.** RMS errors and training times of MLFN models with the change of nodes.

a 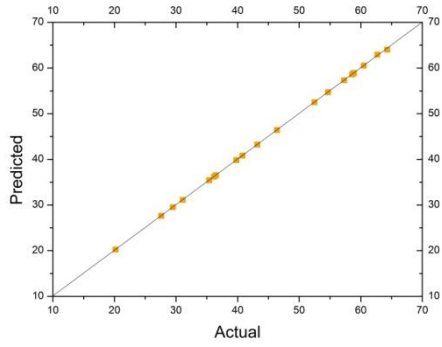 b 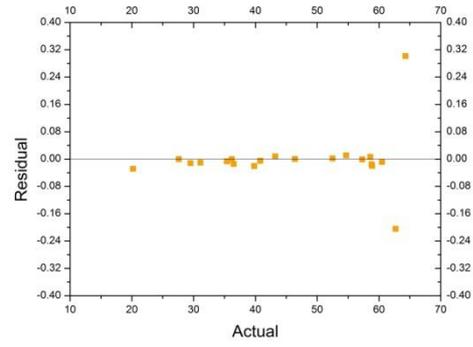

c 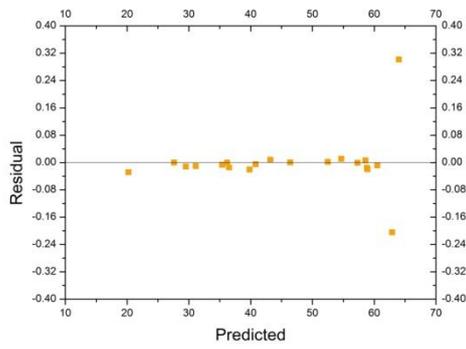

**Figure 9.** Training results of the GRNN model. a) Predicted values versus actual values; b) residual values versus actual values; c) residual values versus predicted values.

a 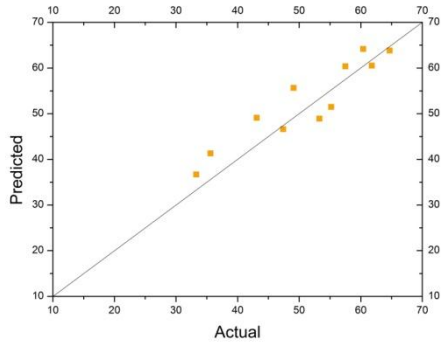  b 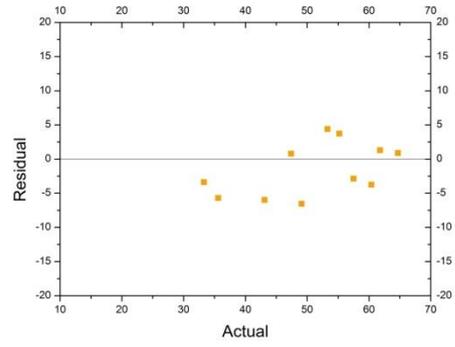

c 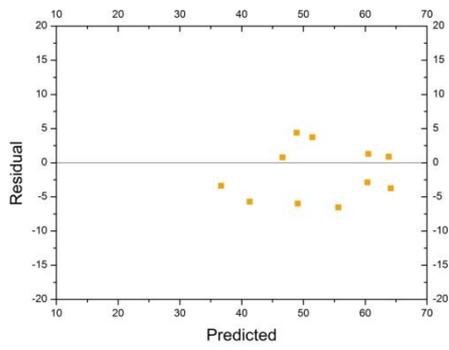

**Figure 10.** Testing results of the GRNN model. a) Predicted values versus actual values; b) residual values versus actual values; c) residual values versus predicted values.

a 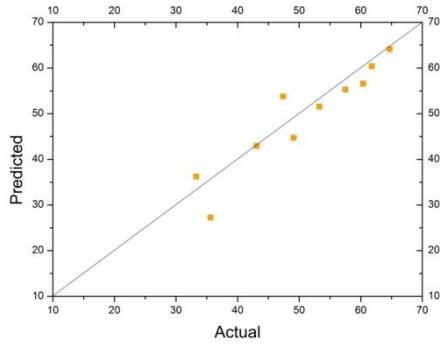 b 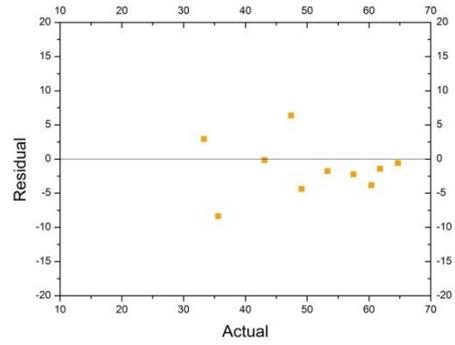

c 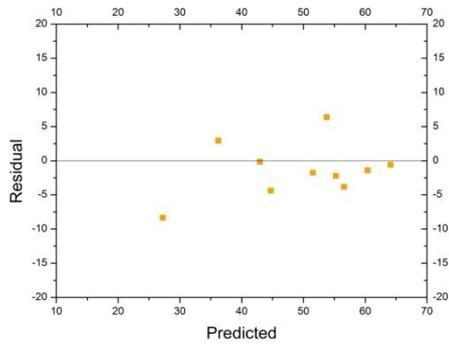

**Figure 11.** Testing results of the SVM model. a) Predicted values versus actual values; b) residual values versus actual values; c) residual values versus predicted values.